\documentclass[letterpaper]{article} % DO NOT CHANGE THIS
\usepackage{aaai23}  % DO NOT CHANGE THIS
\usepackage{times}  % DO NOT CHANGE THIS
\usepackage{helvet}  % DO NOT CHANGE THIS
\usepackage{courier}  % DO NOT CHANGE THIS
\usepackage[hyphens]{url}  % DO NOT CHANGE THIS
\usepackage{graphicx} % DO NOT CHANGE THIS
\def\UrlFont{\rm}  % DO NOT CHANGE THIS
\usepackage{natbib}  % DO NOT CHANGE THIS AND DO NOT ADD ANY OPTIONS TO IT
\usepackage{caption} % DO NOT CHANGE THIS AND DO NOT ADD ANY OPTIONS TO IT
\usepackage{algorithm}
\usepackage{algorithmic}

\usepackage{newfloat}
\usepackage{listings}
\usepackage{algorithm}
\usepackage{algorithmic}
\usepackage{color,xcolor}
\usepackage{amsmath}
\usepackage{subfigure}

\usepackage{array}
\usepackage{multirow}

\usepackage{graphicx}
\usepackage{epsfig}

\usepackage{caption}
\usepackage{listings}
\usepackage{subfloat}
\usepackage{adjustbox}
\usepackage{pifont}
\usepackage{threeparttable}

\usepackage{amssymb}
\usepackage{bbding}

\usepackage{helvet} % DO NOT CHANGE THIS
\usepackage{courier}  % DO NOT CHANGE THIS
\usepackage[hyphens]{url}  % DO NOT CHANGE THIS
\usepackage{graphicx} % DO NOT CHANGE THIS
\usepackage{algorithm}
\usepackage{algorithmic}
\usepackage{color,xcolor}
\usepackage{array}
\usepackage{multirow}

\usepackage{amsmath}

\usepackage{amssymb}

\DeclareCaptionStyle{ruled}{labelfont=normalfont,labelsep=colon,strut=off} % DO NOT CHANGE THIS
\floatstyle{ruled}
\newfloat{listing}{tb}{lst}{}
\floatname{listing}{Listing}
\newcommand{\ignore}[1]{}

\title{A Novel Membership Inference Attack against Dynamic Neural Networks by Utilizing Policy Networks Information}

\usepackage{bibentry}
\begin{document}

\author{Pan Li$^{1,2}$, Peizhuo Lv$^{1,2}$, Shenchen Zhu$^{1,2}$, Ruigang Liang$^{1,2}$, Kai Chen$^{1,2}$ \\ 
\normalsize{$^{1}$SKLOIS, Institute of Information Engineering, Chinese Academy of Sciences, China} \\ 
\normalsize{$^{2}$School of Cyber Security, University of Chinese Academy of Sciences, China} \\ 
\{lvpeizhuo, lipan, zhushenchen\}@iie.ac.cn, shengzhi@bu.edu, \{chenkai, liangruigang\}@iie.ac.cn}

\maketitle

\begin{abstract}
Unlike traditional static deep neural networks (DNNs), dynamic neural networks (NNs) adjust their structures or parameters to different inputs to guarantee accuracy and computational efficiency. Meanwhile, it has been an emerging research area in deep learning recently. Although traditional static DNNs are vulnerable to the membership inference attack (MIA) , which aims to infer whether a particular point was used to train the model, little is known about how such an attack performs on the dynamic NNs. In this paper, we propose a novel MI attack against dynamic NNs, leveraging the unique policy networks mechanism of dynamic NNs to increase the effectiveness of membership inference. We conducted extensive experiments using two dynamic NNs, i.e., GaterNet, BlockDrop, on four mainstream image classification tasks, i.e., CIFAR-10, CIFAR-100, STL-10, and GTSRB. The evaluation results demonstrate that the control-flow information can significantly promote the MIA. Based on backbone-finetuning and information-fusion, our method achieves better results than baseline attack and traditional attack using intermediate information.
\end{abstract}

\section{Introduction}

Dynamic neural networks (NNs) are an emerging technology that can adjust their structures or parameters according to the input during inference and has notable advantages, e.g., improved representation power, computational efficiency, etc. Although traditional static NNs used to be dominant in the domain of deep learning~\cite{Goodfellow2015DeepLearning,deepLearning1,deepLearning2,deepLearning3,deepLearning4}, as a general and substitute approach, dynamic models are also with a wide range of applications, such as image classification~\cite{ImageClassification,imageClassification1,imageClassification2}, object detection~\cite{ObjectDetection,objectDetection1,objectDetection2}, and semantic segmentation~\cite{semanticSegmentation,semanticSegmentation1,semanticSegmentation2}, due to their comparable or superior performance. There are many ways to realize a dynamic neural network, and the way based on a policy network is mainstream. The policy network based dynamic NNs (PN-DNN) includes two independent parts: main network and policy network. The main network aims to identify the target samples, while the policy network is responsible for guiding the identification process of the main network. By introducing the policy network to realize the dynamic structure and parameters, such kind of dynamic neural networks always show good performance.

Despite the success of dynamic NNs in computer vision (CV) and natural language processing (NLP) areas, which are known to be vulnerable to adversarial attacks. For example, DeepSloth~\cite{deepSloth} increases the average computational cost of the dynamic NNs by modifying the objective function, thereby amplifying the latency of the IoT devices. RANet~\cite{RANet} allows each input to adaptively choose an exit layer, increasing the variations and flexibility of adversarial examples. However, whether dynamic NNs are vulnerable to membership inference attacks (MIA) is still unknown. In MIA, the attacker can determine whether a specific record is included or not in the training data.

In fact, membership inference attacks have already been widely researched in traditional static NNs. 
%\lp{According to the adversarial knowledge, membership inference attacks can be divided into black-box and white-box attacks. The black-box attacks[XX,XX,XX] only use the training data distribution and query results on the target model. The white-box attacks[XX,XX,XX] can additionally collect the structure and parameters of the model as well as the intermediate information generated in the running process.}
They can be roughly classified into two categories: binary classifier based attacks and metrics based attacks. The main goal of binary classifier based attacks is to train a binary-classification attack model and use it to distinguish between members and non-members. For example, Shokri et al.~\cite{firstMIA} proposed to train multiple shadow models to mimic the behavior of the target model, and then build up an attack model based on the shadow models for MIAs. In contrast, metrics based attacks~\cite{mlLeak_metricsBased,Yeom2018PrivacyRI_metricsBased} do not require an auxiliary model. They directly rely on the inherent attributes of the target model for discrimination, such as correctness, loss, and entropy. Metrics based attacks have less computational cost while binary classifier based attacks achieve better attack effect. 
%In addition, the MIA on static NNs has been systematically studied. 
%\lpz{I think you should not introduce the related work in a black-box/white-box manner. In this case, the reviewer will overthink that our attack is a white-box approach which may not be as good as the black-box approach. Just introduce some classic papers.}

In this paper, we first investigate the membership inference attack on dynamic NNs using policy networks and find that the control-flow information of policy networks can be exploited to improve the performance of MIAs further. Based on this discovery, we propose a novel MIA on dynamic NNs using the control-flow information of policy networks, which consists of two core phases. In the first phase, we train some shadow models by fine-tuning the target model on an alternative training dataset to obtain combined information of alternative training samples and non-training samples, including the policy's control-flow information network and the output confidence scores of the dynamic NN.
In the second phase, we use the combined information of alternative training samples (i.e., marked as members) and alternative test samples (i.e., marked as non-members) to train an attack model to predict whether a point is in the training dataset of the target model.

We conduct our evaluation on two state-of-the-art dynamic NNs (GaterNet and BlockDrop), two benchmark models (ResNet and VGG), and four mainstream image classification tasks (CIFAR-10, CIFAR-100, STL-10, and GTSRB). The experimental results prove that our proposed attack outperforms the baseline attack and achieves state-of-the-art performance, e.g., for ResNet-based GaterNet trained on CIFAR-100 with 30,000 labeled samples, our attack achieves 72.43\% attack success rate(ASR) while the corresponding baseline attack only  achieve 66.64\%. It indicates that our attack can better reveal the privacy information in the dynamic neural network by utilizing the policy network information (i.e., control-flow information). Moreover, we find that the control-flow information is a more distinctive feature that can effectively be utilized to distinguish the distribution difference between the training and test data. For example, we use 336 dimensional control-flow information to assist the classification of attack models, and the attack success rate is higher than using 4,096 dimensional gradient and activation information (i.e., non-control flow information).

\vspace{3pt}\noindent\textbf{Contributions}. Our main contributions are outlined below:

\vspace{1pt}\noindent$\bullet$\space We propose an attack method based on finetuning and control-flow information, achieving state-of-the-art attack performance. To our best knowledge, this is the first study of the privacy disclosure in dynamic NNs based on policy networks.  

\vspace{1pt}\noindent$\bullet$\space We propose a new perspective on how to effectively implement membership inference attacks, and perform quantitative and qualitative analyses to demonstrate the effectiveness of control-flow information. 

\vspace{1pt}\noindent$\bullet$\space We evaluate the classic defense against our control-flow based attacks and illustrate that our novel attacks can still achieve reasonable performance.

\section{Related Work}

\subsection{Dynamic Neural Networks}
Dynamic neural networks can dynamically adjust structures and parameters according to input samples. Thus, using different computational graphs for simple and complex samples, dynamic NNs achieve higher accuracy, computational efficiency, etc. According to the inference strategy, we can divide dynamic NNs into three main categories: policy networks based, confidence based, and gating functions based.

\vspace{2pt}\noindent$\bullet$\space\textit{Policy Networks-based.} Policy network is an additional network used to learn a decision function for executing multiple units in a model. For example, BlockDrop~\cite{Wu2018BlockDropDI} generates binary vectors through a policy network to determine whether each layer of the dynamic NN participates in the calculation. GaterNet~\cite{gaterNet} also dynamically selects the structure and parameters of the dynamic NNs according to the output of the policy network, and it can reach the channel level. Dynamic neural networks based on policy networks usually have excellent performance and are widely used. Therefore, this paper mainly focuses on member inference attacks in this field.

\vspace{2pt}\noindent$\bullet$\space\textit{Confidence-based.}
Confidence-Based dynamic NNs make decisions based on the confidence of intermediate predictions. If the target confidence reaches a predefined threshold, the model terminates in advance and outputs. For example, MSDNets~\cite{msdnet} allow output in any intermediate layer, discarding the rest of the model structure in depth. It decides by comparing the confidence of intermediate predictions with a predefined threshold. Resolution Adaptive Networks~\cite{RANet} decompose the samples into features with different resolutions and then decide whether to adopt the intermediate layer results depending on the prediction's confidence.

\vspace{2pt}\noindent$\bullet$\space\textit{Gate function-based.}
Gate function-based dynamic NNs insert many binary classifiers in the model. During inference, each plug-in module controls whether to skip the corresponding layer or block. For example, DynamicRouting~\cite{Sabour2017DynamicRouting} dynamically allocates different paths to samples with the help of intermediate classifiers. NeuralPruning~\cite{Lin2017RuntimeNeuralPruning} uses the binary vector generated by the gate functions to determine whether each channel needs to be activated. Gating functions usually have good generality and applicability. However, because they are nondifferentiable, these gating functions usually need specific training techniques.

\subsection{Membership Inference Attacks}
Shokri et al.~\cite{firstMIA} first proposed the concept of Membership Inference Attack. They train a large number of shadow models to learn the distribution difference between the training data and the testing data and then build an inference model to identify whether a target input belongs to the training dataset. This method is effective, but it requires datasets with the same distribution, and the computational cost of training shadow models is relatively large. Yeom et al.~\cite{Yeom2018PrivacyRI_metricsBased} proposed using the model's inherent metrics, such as prediction correctness and prediction loss, to execute member inference attacks, thus successfully solving the above mentioned problems. Compared with the black-box scenario where only the output logits can be obtained, Nasr et al.~\cite{Nasr_intermediate_gradient_activation} additionally used gradient and activation information to enhance the effect of member inference attacks. In addition, there is a lot of work on specific model structures and application scenarios. For example, LOGAN~\cite{Hayes2019LOGANMI} is a member inference attack targeted explicitly at generative models, and Label-Only Exposures~\cite{labelOnly} studies the possibility of privacy disclosure when only the hard label can be obtained. Though there is comprehensive and systematic research work on membership inference attacks, it is still unknown whether dynamic NNs are vulnerable to MIAs.

% target model: the model that we consider to attack.
% attack model: the binary model used to predict whether a sample is a membership
\section{Attack Approach}
\label{sec:attack-approach}

\begin{figure*}[h]
\centering
\epsfig{figure=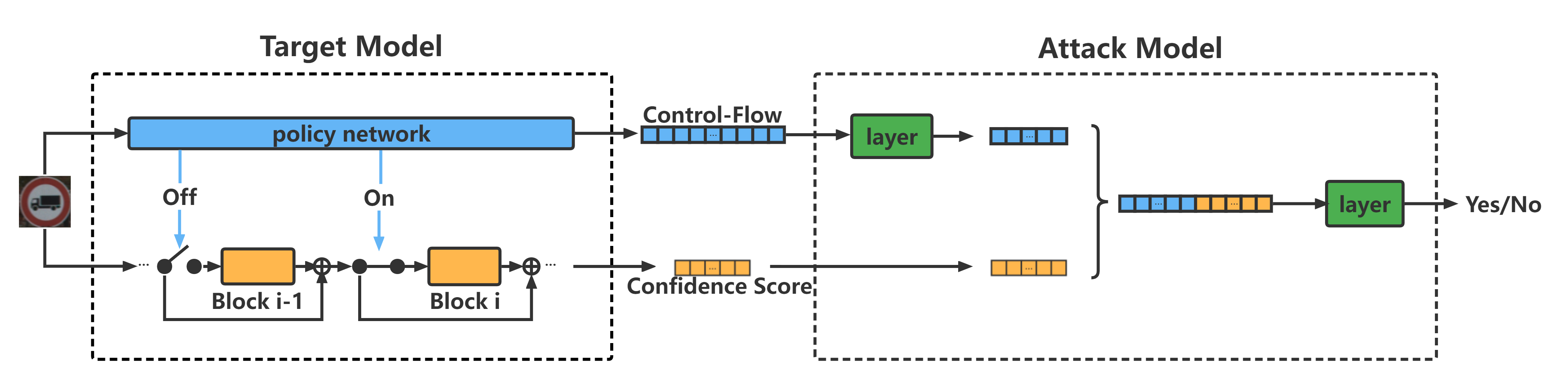, width=\textwidth} 
\caption{Overview of Our Attack. }
\label{fig:workflow}
%\vspace{-4pt}
\end{figure*} 

We propose a membership inference attack against emerging dynamic neural networks. We first describe the threat model and workflow of the proposed attack briefly. Then we introduce the shadow models generation and attack models generation and global algorithm of our attack approach in detail.

\subsection{Threat Model}
Given a target dynamic neural network model, we argue that an attacker can access the target model in a white-box manner. The white-box approach assumes that the attacker can access the ``internals'' of the target model. Thus, the attacker can obtain and utilize the control flow information of the policy network in the dynamic neural network to infer whether a particular point was used to train the model.

\subsection{Attack Workflow}
\label{subsec:workflow}

Figure~\ref{fig:workflow} shows the workflow of our attack approach. Firstly, we split the collected substitute dataset into the training and test datasets and then train the shadow models by fine-tuning the target dynamic neural networks using the substitute training datasets. Secondly, we train a binary model as the attack model to predict whether a sample is a membership. Notably, the attack model is trained on the combination information (consisting of the control flow information of the policy network and the outputs' confidence scores of the dynamic NN) of the substitute training samples (i.e., labeled as the membership) and that of the substitute test samples (i.e., labeled as the non-membership). Finally, for a suspect sample, we can input it into the target dynamic NN to obtain the combination information from the target NN, and input the combination information into the attack model to infer whether the sample is a membership.

\subsection{Shadow Models Training}
\label{subsec:shadow models}

Although an attacker can access the target dynamic NN in a white-box manner, he/she has no access to the label of whether a record belongs to the training dataset of the target model. Therefore, we need to use a shadow model to capture the membership status of the data points in the dataset. In other words, the attacker can generate an alternative dataset and use the alternative dataset to train the shadow model, so he/she can obtain the training dataset of whether a record belongs to the shadow model and further use it to train the attack model to infer whether a point is a member of the training target model.

To quickly obtain a shadow model $f_{s}$ that behaves similarly to the target model $f$, we can fine-tune the target model $f$ on the substitute dataset $D_{s}$. There are two main reasons we choose to fine-tune rather than train the shadow model from scratch. First, as a classical transfer learning method, fine-tuning can ensure that the target model and shadow model have a certain degree of similarity so that the attack model learned based on shadow models can be better applied to the target model. Second, compared with training from scratch, fine-tuning often requires less data. This allows us to reduce the need for the same distributed dataset, or train more shadow models with the same substitute dataset. 
%\todo{Pan. Please give some advantages of the fine-tuning process, i.e., why we use the fine-tuning. And I think the introduction can also solve the doubts: (2.why we use fine-tuning not training from scratch.)} 
There are three options for the fine-tuning process: fine-tuning policy network, fine-tuning the main network, fine-tuning the main network and policy network.

\vspace {2pt}\noindent$\bullet$\space\textit{Fine-tuning policy network.} To obtain the shadow model, we freeze the parameters of the main network in the target model and only use the substitute data to fine-tune the policy network. It gives the shadow model the same feature extractor as the target model, but the control flow information will adapt to the distribution of the substitute data.
%\lpz{Introduce the process of it and give an intuition of it (i.e., a straightforward understanding of it).}

\vspace {2pt}\noindent$\bullet$\space\textit{Fine-tuning main network.}
In this setting, we freeze the parameters of the policy network in the target model and only use the substitute data to fine-tune the main network to obtain the shadow model. This gives the shadow model the same controller as the target model, but the feature extractor will adapt to the distribution of substitute data.

\vspace {2pt}\noindent$\bullet$\space\textit{Fine-tuning both policy network and main network.}
In this setting, we optimize the parameters of the main network and the policy network in the target model to obtain a shadow model. The distribution of the substitute data will influence both the feature extractor and the controller, and the feature extractor and the controller will also affect each other.

%\lpz{Introduce which one is better in the evaluation, and why we use it?}
%\lpz{We can give two options and introduce that we should choose fine-tuning because of the randomness of the policy network, and the randomness results in the failure of the attack model.}

Moreover, according to our evaluation, fine-tuning main network is the best choice to train the shadow models and fine-tuning policy network is the worst choice\footnote{We trained multiple GaterNets based on four datasets to evaluate the above three fine-tuning methods. The average attack success rate of the three fine-tuning methods (i.e., Fine-tuning policy network, Fine-tuning main network, and  Fine-tuning both policy network and main network) is 0.6341, 0.8039, and 0.7336, respectively.}. We suspect that this is because the policy network is sensitive to fine-tuning. 
As we know, the output of the policy network is usually a small amount of information (e.g., 336-dimension binary vector), which controls the behavior of the main network. This means that every bit of the policy network information is significant. Even if we fine-tune the policy network slightly, it may seriously influence the policy network's decision. Thus the attack model trained on the shadow model with the policy network tuned cannot be applied to the target model, leading to the failure of our attacks.
Therefore, we finally choose only fine-tuning main network to obtain the shadow model.

% an approximated set of training data is desired to imply membership. This set can be obtained either by:

% The shadow model was used to capture membership status of data points in datasets.

% In particular, shadow model is proposed to imitate target model’s behavior by training on a similar dataset.

% The problem is that the output label of whether a record belongs to the dataset of target model cannot be obtained. So here attackers often generate sub- stituted dataset by data synthesis. The input of this training is generated either by the shadow model. The attack model training process first selects some records from both inside and outside the substituted dataset, and then obtains the class probability vector through target model or shadow model. The vector and the label of record are taken as input, and whether this record belongs to substituted dataset is taken as output.

\subsection{Attack Models Generation}
Given the shadow model $f_{s}$, we aim to design an attack model $f_{a}$ which can utilize the unique control-flow information of the policy network to improve its inference accuracy. Therefore, as shown in Figure~\ref{fig:workflow}, we design the attack model $f_{a}$ which takes both the control flow information and the outputs confidence scores of the target models as the inputting information and processes this two information separately through two NN branches. Then, $f_{a}$ concatenates these two intermediate processed information and computes the concatenated information to infer whether a sample was a training sample. 

For training the attack model $f_{a}$, we first input the samples of the training substitute dataset and the samples of the non-training substitute dataset into the shadow model $f_{s}$. We obtain the control-flow information and output confidence scores information for training substitute samples ($I_{c\_s}$, $I_{o\_s}$) and that of the non-training substitute samples ($I_{c\_ns}$, $I_{o\_ns}$).
Then we label the information ($I_{c\_s}$, $I_{o\_s}$) of training substitute samples as ``membership'', and the information ($I_{c\_ns}$, $I_{o\_ns}$) of non-training substitute samples as ``no-membership'', thus, we can obtain the training dataset $D_{a}$ of the attack model $f_{a}$. And we can train $f_{a}$ using $D_{a}$ as below:
\begin{equation}
\label{loss:attack-model}
f_{a} = \mathop{argmin}_{f_{a}}\sum\nolimits_{(I_{c}, I_{o})_{i}, y_{i} \in D_{a}} \mathcal{L}(f_{a}((I_{c}, I_{o})_{i}), y_{i})
\end{equation} 
Thus, to infer whether a suspect sample $x_{s}$ is a membership of the target model, we can input it into the target model and obtain the control flow information and output confidence scores information ($I_{c\_nt}$, $I_{o\_nt}$). Then we input ($I_{c\_nt}$, $I_{o\_nt}$) into our attack model and obtain the result to show whether the sample is a membership.

\subsection{Global Algorithm}
We summarize our attack method in Algorithm~\ref{alg:Position-injected}. When we get a target model to attack (Line 1-3), we first freeze its policy network (Line 4), and then we fine-tune its main network with the same distribution datasets $D_{train}^{shadow}$ and $D_{test}^{shadow}$ to obtain our shadow model (Line 5-7). Next, we use the shadow model to process $D_{train}^{shadow}$ and $D_{test}^{shadow}$, and obtain the logits and policy network information corresponding to each sample. Further, we label the samples in $D_{train}^{shadow}$ as a member and the samples in $D_{test}^{shadow}$ as a non-member for training the attack model (Line 8-12). Finally, we use the target model to process $D_{train}^{target}$ and $D_{test}^{target}$ to obtain the test dataset, and verify the attack membership inference ASR on the attack model (Line 13-15).

\begin{algorithm}[t]
 	\caption{Global Algorithm}
 	\label{alg:Position-injected}
 	\begin{algorithmic}[1]
 	\REQUIRE $f_t$: The target model, include $ft\_main$ and $ft\_policy$; $f_s$: The shadow model, include $fs\_main$ and $fs\_policy$; $f_a$: The attack model.
 	\ENSURE $ASR$: Attack Success Rate 
	\FOR{$i$ steps}                        
	\STATE{$f_t$ = TRAIN($D_{train}^{target}$, $D_{test}^{target}$)}
	\ENDFOR

    \STATE{Freeze the parameters of $ft\_policy$}
    \FOR{$j$ steps}
	\STATE{$f_s$ = FINETUNE($ft$, $D_{train}^{target}$, $D_{test}^{target}$)}
    \ENDFOR

    \STATE{$D_{member}^{shadow}$ = [$f_s$($D_{train}^{shadow}$),$fs\_policy$($D_{train}^{shadow}$),$\overrightarrow{1}$]}
    
    \STATE{$D_{non-member}^{shadow}$ = [$f_s$($D_{test}^{shadow}$),$fs\_policy$($D_{test}^{shadow}$),$\overrightarrow{0}$]}

    \FOR{$k$ steps}    \STATE{$f_a$=TRAIN($D_{member}^{shadow}$,$D_{non-member}^{shadow}$)}
    \ENDFOR

    \STATE{$D_{member}^{target}$ = [$f_t$($D_{train}^{target}$),$ft\_policy$($D_{train}^{target}$),$\overrightarrow{1}$]}
    
    \STATE{$D_{non-member}^{target}$ = [$f_t$($D_{test}^{target}$),$ft\_policy$($D_{test}^{target}$),$\overrightarrow{0}$]}
    \STATE{$ASR$=TEST($f_a$, $D_{member}^{target}$, $D_{non-member}^{target}$)}
    
    \RETURN $ASR$
\end{algorithmic}
\end{algorithm}

% For a model F and its training dataset D, training attack model needs information of label x, F(x), and whether x $\in$ D. If using a shadow model, shadow model F and its dataset D are known. All information is from shadow model and corresponding dataset. If using the target model, F is the target model and D is the training dataset. However, attackers do not know D.

% \lpz{1. what is attack model and how to design the attack model (mainly introduce the combination information.)?}

% \lpz{2. how to train the attack model?}

% \lpz{Architecture; inputs}

%\vspace{-6pt}
\section{Experiment}
\subsection{Experiment Settings}

\begin{table*}[h]
\centering
\caption{The Performance of Our Attack}
\label{tab:Performance}
\begin{tabular}{c|c|c|c|c|c|c|c} 
\hline
\begin{tabular}[c]{@{}c@{}}\textbf{Dynamic}\\\textbf{Neural Network}\end{tabular} & \begin{tabular}[c]{@{}c@{}}\textbf{Model}\\\textbf{Structure}\end{tabular} & \textbf{Dataset} & \begin{tabular}[c]{@{}c@{}}\textbf{Target}\\\textbf{ACC}\end{tabular} & \begin{tabular}[c]{@{}c@{}}\textbf{Shadow}\\\textbf{ACC}\end{tabular} & \textbf{ASR} & \textbf{Recall} & \textbf{Precision} \\ 
\hline
\hline
\multirow{8}{*}{GaterNet} & \multirow{4}{*}{VGG} & CIFAR-10 & 83.82\% & 94.11\% & 52.33\% & 0.9918 & 0.5120 \\
 &  & CIFAR-100 & 33.47\% & 40.84\% & 54.50\% & 0.4999 & 0.5495 \\
 &  & STL10 & 78.97\% & 79.03\% & 65.11\% & 0.8118 & 0.6144 \\
 &  & GTSRB & 96.99\% & 98.06\% & 64.61\% & 0.6353 & 0.6494 \\ 
\cline{2-8}
 & \multirow{4}{*}{Resnet} & CIFAR-10 & 85.17\% & 95.99\% & 66.39\% & 0.5181 & 0.7313 \\
 &  & CIFAR-100 & 46.86\% & 61.67\% & 72.43\% & 0.5962 & 0.8017 \\
 &  & STL10 & 73.40\% & 74.01\% & 82.15\% & 0.6859 & 0.9413 \\
 &  & GTSRB & 99.68\% & 99.97\% & 56.10\% & 0.9300 & 0.5351 \\
\hline

\multirow{8}{*}{BlockDrop} & \multirow{4}{*}{VGG} & CIFAR-10 & 83.46\% & 84.73\% & 72.32\% & 0.7743 & 0.7042 \\
 &  & CIFAR-100 & 42.76\% & 45.84\% & 66.18\% & 0.5525 & 0.8802 \\
 &  & STL10 & 72.28\% & 74.58\% & 68.35\% & 0.7456 & 0.6682 \\
 &  & GTSRB & 97.32\% & 98.24\% & 64.14\% & 0.6226 & 0.7416 \\ 
\cline{2-8}
 & \multirow{4}{*}{Resnet} & CIFAR-10 & 85.48\% & 88.58\% & 79.90\% & 0.8831 & 0.7560 \\
 &  & CIFAR-100 & 55.80\% & 60.04\% & 69.25\% & 0.5905 & 0.9861 \\
 &  & STL10 & 74.46\% & 75.32\% & 80.16\% & 0.6457 & 0.9236 \\
 &  & GTSRB & 98.75\% & 98.83\% & 70.15\% & 0.6846 & 0.8322 \\
\hline

\end{tabular}
\end{table*}

\begin{table*}[h]
\centering
\caption{The Performance of Baseline Attack}
\label{tab:baseline}
\begin{tabular}{c|c|c|c|c|c|c|c} 
\hline
\begin{tabular}[c]{@{}c@{}}\textbf{Dynamic}\\\textbf{Neural Network}\end{tabular} & \begin{tabular}[c]{@{}c@{}}\textbf{Model}\\\textbf{Structure}\end{tabular} & \textbf{Dataset} & \begin{tabular}[c]{@{}c@{}}\textbf{Target}\\\textbf{ACC}\end{tabular} & \begin{tabular}[c]{@{}c@{}}\textbf{Shadow}\\\textbf{ACC}\end{tabular} & \textbf{ASR} & \textbf{Recall} & \textbf{Precision} \\ 
\hline
\hline
\multirow{8}{*}{GaterNet} & \multirow{4}{*}{VGG} & CIFAR-10 & 83.82\% & 83.89\% & 51.41\% & 0.4999 & 0.5146 \\
 &  & CIFAR-100 & 33.47\% & 37.78\% & 50.58\% & 0.4557 & 0.5064 \\
 &  & STL-10 & 78.97\% & 72.37\% & 54.18\% & 0.5690 & 0.5396 \\
 &  & GTSRB & 96.99\% & 99.33\% & 49.57\% & 0.6004 & 0.4965 \\ 
\cline{2-8}
 & \multirow{4}{*}{Resnet} & CIFAR-10 & 85.17\% & 84.75\% & 66.77\% & 0.4786 & 0.7697 \\
 &  & CIFAR-100 & 46.86\% & 49.59\% & 66.64\% & 0.5871 & 0.6977 \\
 &  & STL10 & 73.40\% & 72.38\% & 76.26\% & 0.8648 & 0.7180 \\
 &  & GTSRB & 99.68\% & 99.70\% & 52.82\% & 0.4618 & 0.5325 \\
\hline
\end{tabular}
\end{table*}

\begin{figure*}[htbp]
\centering

\subfigure[CIFAR-10]{
\begin{minipage}[t]{0.22\linewidth}
\centering
\includegraphics[width=1.6in]{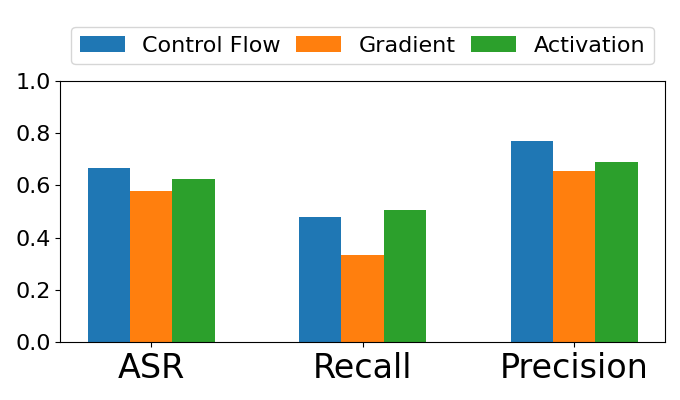}
%\caption{fig1}
\end{minipage}%
}%
\subfigure[CIFAR-100]{
\begin{minipage}[t]{0.22\linewidth}
\centering
\includegraphics[width=1.6in]{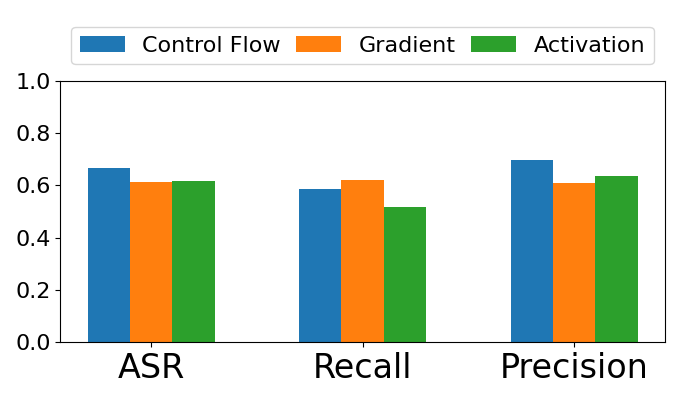}
%\caption{fig2}
\end{minipage}%
}%
\subfigure[STL-10]{
\begin{minipage}[t]{0.22\linewidth}
\centering
\includegraphics[width=1.6in]{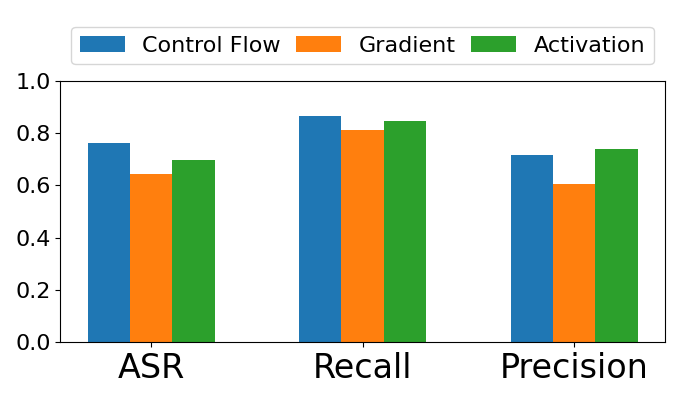}
%\caption{fig1}
\end{minipage}%
}%
\subfigure[GTSRB]{
\begin{minipage}[t]{0.22\linewidth}
\centering
\includegraphics[width=1.6in]{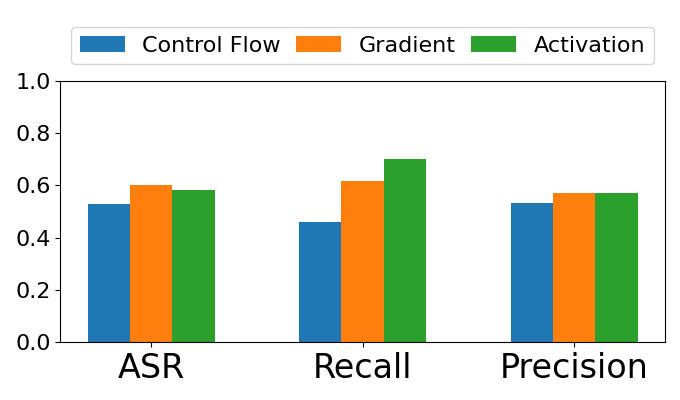}
%\caption{fig2}
\end{minipage}%
}%

\centering
\caption{Comparison of MIAs using Gradient or Activation Information.}
\label{fig:comparison experiment}
\end{figure*}

\vspace{2pt}\noindent\textbf{Policy Network.} 
We consider two popular dynamic NNs based on policy networks, GaterNet and BlockDrop, which are both composed of main network and policy network. In order to evaluate the experimental results more objectively, we use ResNet-18 and VGG to realize the main network of each dynamic NN and follow the source code of GaterNet and BlockDrop to build up the policy network. When we leverage the dynamic neural network to predict, the policy network first processes the input sample to obtain N-dimension control-flow information. Then the main network receives the sample and the control-flow information together as inputs, and the control-flow information guides the main network to predict the specific sample. We use the complete dynamic neural network (main network plus policy network) as our target model. The shadow model shares the same structure and parameters as the target model. Specifically, we apply the Adam optimizer to train the target and shadow models. The initial learning rate was 0.01 and decreased to 0.0001 after 100 epochs by CosineAnnealingLR.

\vspace{2pt}\noindent\textbf{Attack Model.} 
Our attack model consists of two parts. The first part is a 3-layer MLP, which receives the policy network information as input and outputs a 10-dimensional vector. This part mainly plays the role of reducing dimension and removing redundant noise. The second part is also a 3-layer MLP, and we take the logits vector and 10-dimension policy network vector as input. The second part finally outputs a confidence score between 0 and 1, indicating the probability that a sample belongs to the member samples. The main function of the second part is to combine logits information and policy network information to make judgments. We train the attack model for 100 epochs using an Adam optimizer with a learning rate of 0.001 and a batch size of 1,000, and we leverage binary cross entropy as the loss function to supervise the model training.

\vspace{2pt}\noindent\textbf{Datasets.} 
We evaluate the performance of MIAs and defenses on four popular datasets (i.e., CIFAR-10, CIFAR-100, STL-10, and GTSRB). For each dataset, we first combined the training data and the testing data, and then randomly divided them into four parts, i.e., $D_{train}^{target}$, $D_{test}^{target}$, $D_{train}^{shadow}$, and $D_{test}^{shadow}$. We use $D_{train}^{target}$ to train the target model, and mark the samples in $D_{train}^{target}$ as members of the target model. We use $D_{test}^{target}$ to test the target model, and mark the samples in $D_{test}^{target}$ as non-members of the target model. $D_{train}^{shadow}$ and $D_{test}^{shadow}$ are applied to build up the shadow model and the attack model. Note that since the sample size of the original dataset is limited, in order to ensure the performance of the target model and the shadow model, there will be some overlap between $D_{train}^{target}$ and $D_{train}^{shadow}$. Moreover, both $D_{train}^{target}$ and $D_{train}^{shadow}$ are smaller than the original training dataset, which will lead to a certain decrease in the accuracy of the target model.

\vspace{2pt}\noindent\textbf{Metrics.}
As in previous work\cite{He2021QuantifyingAM, firstMIA, Zhang2021FlexMatchBS}, we use the classification accuracy(ACC) on the test dataset to evaluate the model's performance. As for the effect of membership inference attacks, we adopt attack success rate(ASR), precision and recall as evaluation metrics. 

\vspace{1pt}\noindent$\bullet$  \textit{ASR}. When we calculate ASR, we first take the same number of samples from $D_{train}^{target}$ and $D_{test}^{target}$ as members and non-members, respectively. Then we mark members as 1 and non-members as 0 to get the target label of each sample. Finally, we use the attack model to predict all the samples and calculate the accuracy as ASR.

\vspace{1pt}\noindent$\bullet$  \textit{Precision}. True Positive(TP) refers to the number of member samples correctly predicted by the attack model. False Positive(FP) refers to the number of member samples mistakenly predicted by the attack model. The evaluation metric Precision can be formalized as:
\begin{equation}
\label{precision}
Precision = \frac{TP}{TP+FP} 
\end{equation}

\vspace{1pt}\noindent$\bullet$  \textit{Recall}. True Positive(TP) refers to the number of member samples correctly predicted by the attack model. False Negative(FN) refers to the number of non-member samples mistakenly predicted by the attack model. The evaluation metric Recall can be formalized as:
\begin{equation}
\label{recall}
Recall = \frac{TP}{TP+FN} 
\end{equation}

\vspace{2pt}\noindent\textbf{Platform}. All our experiments are conducted on a server running 64-bit Ubuntu 18.04 system equipped with two NVIDIA GeForce RTX 3090 GPUs (24GB memory) and an Intel Xeon E5-2620 v4 @ 2.10GHz CPU, 128GB memory.

\subsection{Model Performance}
We first evaluate the accuracy of the target model and shadow model on the original task. We use $D_{test}^{target}$ to test the target model and $D_{test}^{shadow}$ to test the shadow model. The results in Table~\ref{tab:Performance} show that whether we use GaterNet or BlockDrop, and whether the main task is based on ResNet or VGG, the original classification task has achieved relatively good performance. In addition, both $D_{train}^{target}$ and $D_{train}^{shadow}$ only contain 30,000 samples, which are part of the original training dataset. This means that more data can further improve the performance of the model. We can also note that the accuracy of simple tasks is higher than that of complex tasks. For example, when the target model is GaterNet based on ResNet, the classification accuracy on simple tasks(CIFAR-10, GTSRB) are 85.17\% and 99.68\%, while the classification accuracy on complex tasks(STL-10, CIFAR-100) are 73.40\% and 46.86\%. This indicates that for complex tasks, we may need to provide more data to ensure the model performance. We can also see that the Shadow ACC is always higher than its corresponding Target ACC. Our shadow model is obtained by fine-tuning the target model using $D_{train}^{shadow}$. Compared with the original training dataset of the target model, $D_{train}^{shadow}$ contains some new data, thus improving the model performance. In addition, the accuracy of ResNet-Based model is higher than that of VGG-Based model, which indicates that more complicated architecture is helpful to fit the training data.

\subsection{Membership Inference Attack Performance}

We also evaluate the performance of our attack and compare the effect with the baseline. Shokri et al.~\cite{firstMIA} proposed the first membership inference attack based on the attack model. It uses $D_{train}^{shadow}$ to build the shadow model from scratch, and inputs the logits information to train the attack model for prediction. We implement this method as the baseline attack for comparison. The result in Table~\ref{tab:baseline} that our method's attack success rate(ASR) is superior to baseline attack in most cases. As for precision and recall, our attack can achieve at least one metric close and the other better, or both metrics better than the baseline attack. For example,  we train ResNet-Based GaterNet with GTSRB dataset, the Precision and Recall of baseline attacks are 0.5325 and 0.4618, and the precision and recall of our method are 0.5351 and 0.9300. Among them, the precision value is close, while the recall of our method is much higher than the baseline attack. The result indicates that our method has higher coverage in detecting member samples. In addition, we can also note that the attack effect of dynamic NNs based on ResNet is better than that of dynamic NNs based on VGG. Moreover, the higher ACC of the target model, the higher ASR of our method, indicating that when the model has a larger capacity, it can better fit the training data and improve the effect of member inference attacks.

\begin{figure*}[h]
\centering
\subfigure[CIFAR-10]{
\begin{minipage}[t]{0.22\linewidth}
\centering
\includegraphics[width=1.6in]{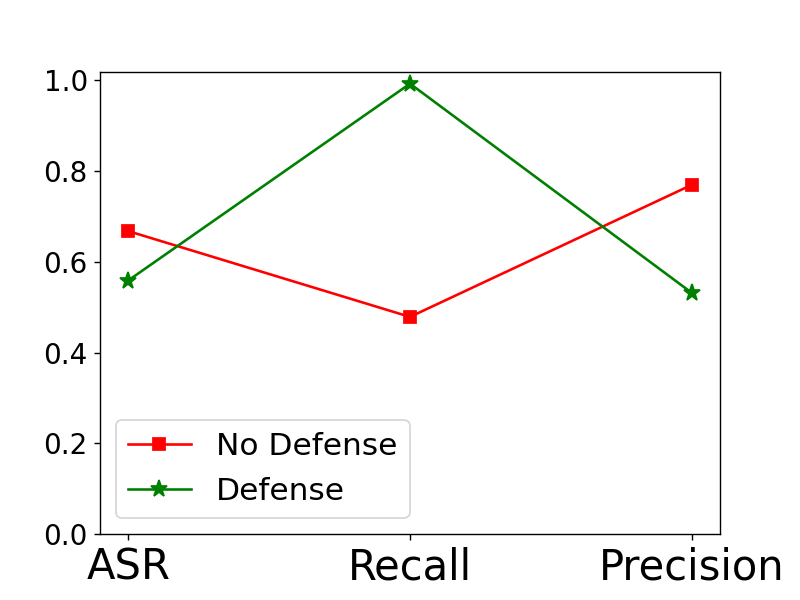}
%\caption{fig1}
\end{minipage}%
}%
\subfigure[CIFAR-100]{
\begin{minipage}[t]{0.22\linewidth}
\centering
\includegraphics[width=1.6in]{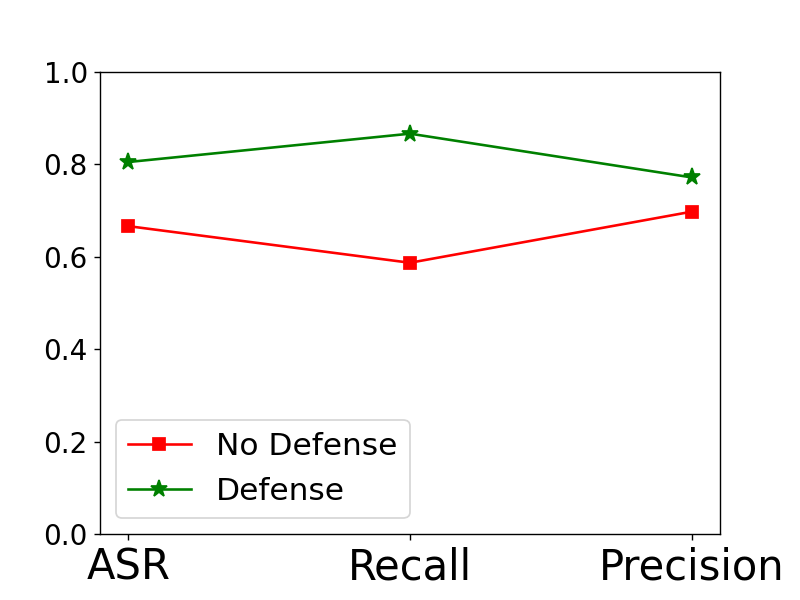}
%\caption{fig2}
\end{minipage}%
}%
\subfigure[STL-10]{
\begin{minipage}[t]{0.22\linewidth}
\centering
\includegraphics[width=1.6in]{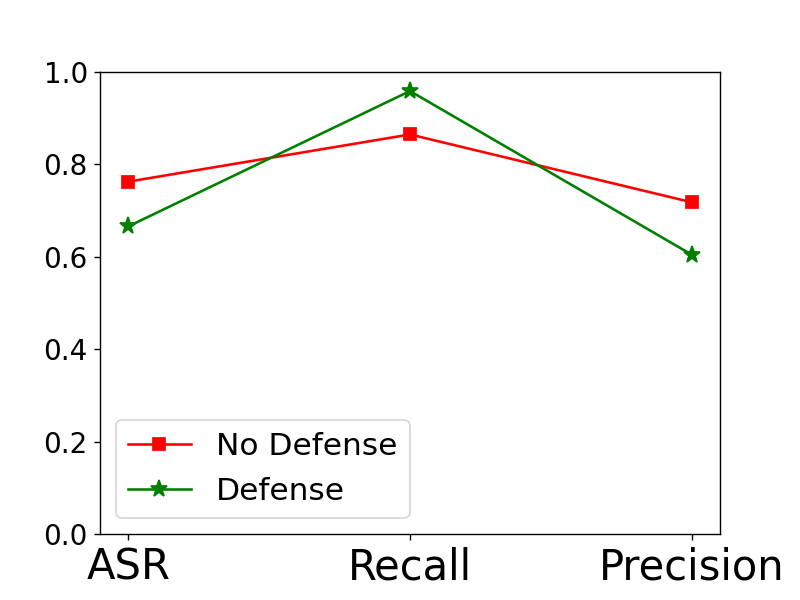}
%\caption{fig1}
\end{minipage}%
}%
\subfigure[GTSRB]{
\begin{minipage}[t]{0.22\linewidth}
\centering
\includegraphics[width=1.6in]{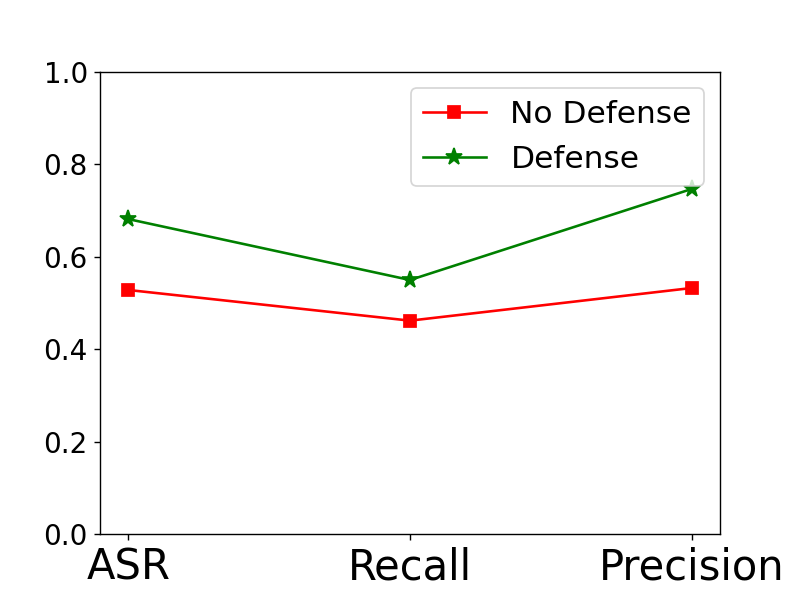}
%\caption{fig2}
\end{minipage}%
}%
\centering
\caption{Our Attack Against Defense Model and Non-Defense Model.}
\label{fig:Against-Defense-Method}
\end{figure*}

\subsection{Comparative Experiment}

We conduct a comparative experiment to verify the importance of policy network information for membership inference attacks. Inspired by Comprehensive Privacy Analysis\cite{Nasr_intermediate_gradient_activation}, we consider using intermediate information such as gradient and activation of the model to execute membership inference attacks. In the comparative experiment, we use the same parameters and network structures as our method, and only replace the policy network information with gradient or activation to observe the difference in results. It should be specially pointed out that the policy network information we use in GaterNet is a 336 dimension vector, while the gradient and activation vectors have 4096 dimensions, which is much larger than the data capacity of the policy network information. In addition, We extract gradient and activation information from the last convolution layer of the main network. Because previous studies~\cite{activationCluster_convLastLayer,trojanAttack_convLastLayer} show that the gradient and activation there contain more high-level semantic information and are closely related to the final classification results. As shown in Figure~\ref{fig:comparison experiment}, the effect of membership inference attack using policy network information is better than that using gradient or activation in most cases. This shows that although the gradient or activation vector has a higher dimension, it does not bring more effective information. Compared with gradient and activation, policy network information is a distinctive feature for membership inference attacks.

\subsection{Against Defense Method}
We implement a classical membership inference defense method Adversarial Regularization~\cite{adversarialRegularization}, and evaluate the performance of the reinforced target model when facing our attack. The core idea of this defense method is to introduce an inference model in the training process of the target model. Inference model and target model should play games with each other. The inference model tries to learn the difference between the member and non-member samples, while the target model should eliminate the difference as much as possible while ensuring the accuracy of the main task. The target model and inference model are trained alternately. When the target model converges, it can defend against membership inference attacks. We first use this defense method to train the main network, and then fine-tune the main network and the policy network together to obtain the target model. Finally, we use our attack method to test the reinforced model. As shown in Figure~\ref{fig:Against-Defense-Method}, the ASR of our attack method has increased on CIFAR-100 and decreased on the other three datasets, but all of them still preserve membership inference ability. The experimental results prove that the defense method has some defense effects, but our attack also shows certain robustness.

\section{Discussion}

In this paper, while we initially demonstrate the importance of policy network information for member inference attacks, some unknown details can still be explored in future directions. On the one hand, the dimensionality of policy network information is an interesting variable. In this paper, we follow the source code of GaterNet and use a 336-dimensional vector. This dimensionality is moderate and brings good results. However, if the dimensionality of the policy network information is very small, can it still provide enough information? Conversely, if the dimensionality of the policy network is enormous, will it introduce noise and affect the effectiveness of the attack?
On the other hand, the relationship between the main and policy networks is an exciting topic. In the PN-DNN scenario, the policy network can directly dictate the structure or parameters of the main network and play a decisive role in the prediction, which leads to the output of the policy network containing a large amount of information. However, if the relationship between the policy network and the main network is not decisive, such as in some interpretable networks where the auxiliary network only plays a role in explaining the prediction results, can the information from the policy network be used for member inference? These exciting questions bring great scope for further investigation of membership inference based on policy networks.
\section{Conclusion}

In this paper, we perform the first membership inference attack against dynamic NNs based on policy networks. We train the shadow model by finetuning the target model, and combine the policy network information to assist the judgement of the attack model. Basic experiments have proved the effectiveness of our attack, which can outperform the baseline attacks in most cases. We also prove the importance of policy network information through comparative experiments. Although using a lower dimension vector, the policy network information can bring a better attack effect than using gradient and activation information. Finally, we implement a classic defense method Adversarial Regularization. Our attack shows sufficient attack effect in the face of the model strengthened by the defense method, which proves the powerful attack effect of our method.

\bibliography{aaai23.bib}

\end{document}